\documentclass[sigconf]{acmart}
\usepackage{color,colortbl}
\usepackage{xcolor,xspace}
\usepackage{paralist, tabularx}
\usepackage{algorithm}
\usepackage{algpseudocode}
\usepackage{multirow}
\usepackage{adjustbox}
\usepackage{xtab}
\usepackage{multicol}
\usepackage{subcaption}
\usepackage{makecell}

\usepackage{longtable}
\usepackage{arydshln}

\newcommand{\xmark}{\text{\sffamily X}} 

\newcommand*{\Scale}[2][4]{\scalebox{#1}{$#2$}}

\AtBeginDocument{%
  \providecommand\BibTeX{{%
    \normalfont B\kern-0.5em{\scshape i\kern-0.25em b}\kern-0.8em\TeX}}}

\copyrightyear{2024}
\acmYear{2024}
\setcopyright{acmlicensed}\acmConference[CIKM '24]{Proceedings of the 33rd ACM International Conference on Information and Knowledge Management}{October 21--25, 2024}{Boise, ID, USA}
\acmBooktitle{Proceedings of the 33rd ACM International Conference on Information and Knowledge Management (CIKM '24), October 21--25, 2024, Boise, ID, USA}
\acmDOI{10.1145/3627673.3679608}
\acmISBN{979-8-4007-0436-9/24/10}

\begin{document}


\title{Chain-of-Layer: Iteratively Prompting Large Language Models for Taxonomy Induction from Limited Examples}


\author{Qingkai Zeng}
\authornote{Equal contribution.}
\email{qzeng@nd.edu}
\affiliation{%
  \institution{University of Notre Dame}
  \city{Notre Dame}
  \state{IN}
  \country{USA}}

\author{Yuyang Bai}
\authornotemark[1]
\authornote{Work done during the visiting student program at the University of Notre Dame.}
\email{ybai3@nd.edu}
\affiliation{%
  \institution{University of Notre Dame}
  \city{Notre Dame}
  \state{IN}
  \country{USA}}

\author{Zhaoxuan Tan}
\email{ztan3@nd.edu}
\affiliation{%
  \institution{University of Notre Dame}
  \city{Notre Dame}
  \state{IN}
  \country{USA}}

\author{Shangbin Feng}
\email{shangbin@cs.washington.edu}
\affiliation{%
  \institution{University of Washington}
  \city{Seattle}
  \state{WA}
  \country{USA}}

\author{Zhenwen Liang}
\email{zliang6@nd.edu}
\affiliation{%
  \institution{University of Notre Dame}
  \city{Notre Dame}
  \state{IN}
  \country{USA}}

\author{Zhihan Zhang}
\email{zzhang23@nd.edu}
\affiliation{%
  \institution{University of Notre Dame}
  \city{Notre Dame}
  \state{IN}
  \country{USA}}

\author{Meng Jiang}
\email{mjiang2@nd.edu}
\affiliation{%
  \institution{University of Notre Dame}
  \city{Notre Dame}
  \state{IN}
  \country{USA}}




\renewcommand{\shortauthors}{Zeng and Bai \emph{et al.}}


\begin{abstract}
Automatic taxonomy induction is crucial for web search, recommendation systems, and question answering. Manual curation of taxonomies is expensive in terms of human effort, making automatic taxonomy construction highly desirable. In this work, we introduce \textsc{Chain-of-Layer} which is an in-context learning framework designed to induct taxonomies from a given set of entities. \textsc{Chain-of-Layer} breaks down the task into selecting relevant candidate entities in each layer and gradually building the taxonomy from top to bottom. To minimize errors, we introduce the Ensemble-based Ranking Filter to reduce the hallucinated content generated at each iteration. Through extensive experiments, we demonstrate that \textsc{Chain-of-Layer}\footnote{The code is available at \url{https://github.com/qingkaizeng/chain-of-layer}.} achieves state-of-the-art performance on four real-world benchmarks.

\end{abstract}


\maketitle

\section{Introduction}
Taxonomy refers to a hierarchical structure that outlines the connections between concepts or entities. It commonly represents these relationships through hypernym-hyponym associations or ``is-a'' relationships. Taxonomies are essential in aiding several tasks, such as textual content understanding~\cite{karamanolakis2020txtract, hao2019universal, xiang2021ontoea}, personalized recommendations~\cite{zhang2014taxonomy, huang2019taxonomy, tan2022enhancing}, and questions answering~\cite{yang2017efficiently}. However, developing a taxonomy solely based on human experts can be a time-consuming and costly process, often presenting challenges in terms of scalability. Consequently, recent efforts have focused on automatic taxonomy induction, which aims to autonomously organize a group of entities into a taxonomy.

\begin{figure}
\centering

\begin{subfigure}[b]{0.45\textwidth}
   \includegraphics[width=\textwidth]{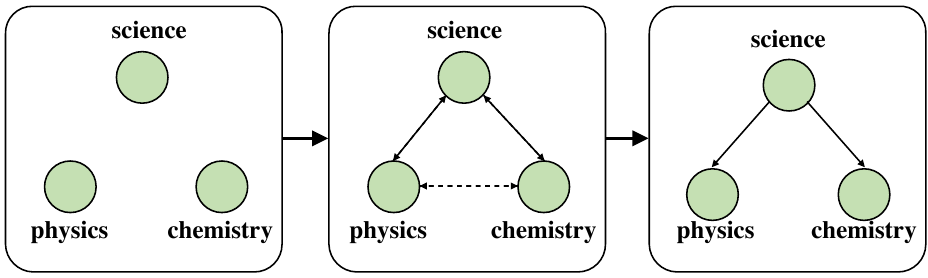}
   \caption{Discriminative Methods: Scoring each entity pair and pruning to taxonomic structure~\cite{shang2020taxonomy,chen2021constructing}}
   \label{fig:motivation_sub1}
\end{subfigure}
\begin{subfigure}[b]{0.45\textwidth}
   \includegraphics[width=\textwidth]{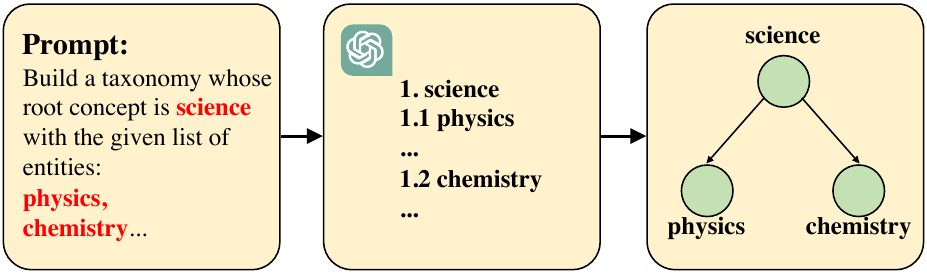}
   \caption{Generative Methods: Prompting LLMs to generate taxonomy}
   \label{fig:motivation_sub2}
\end{subfigure}
\vspace{-0.1in}
\caption{Two Types of Methods for Taxonomy Induction}
\vspace{-0.2in}
\label{fig:motivation}
\end{figure}

\begin{figure*}[t]
    \centering
    \includegraphics[width=1.0\linewidth]{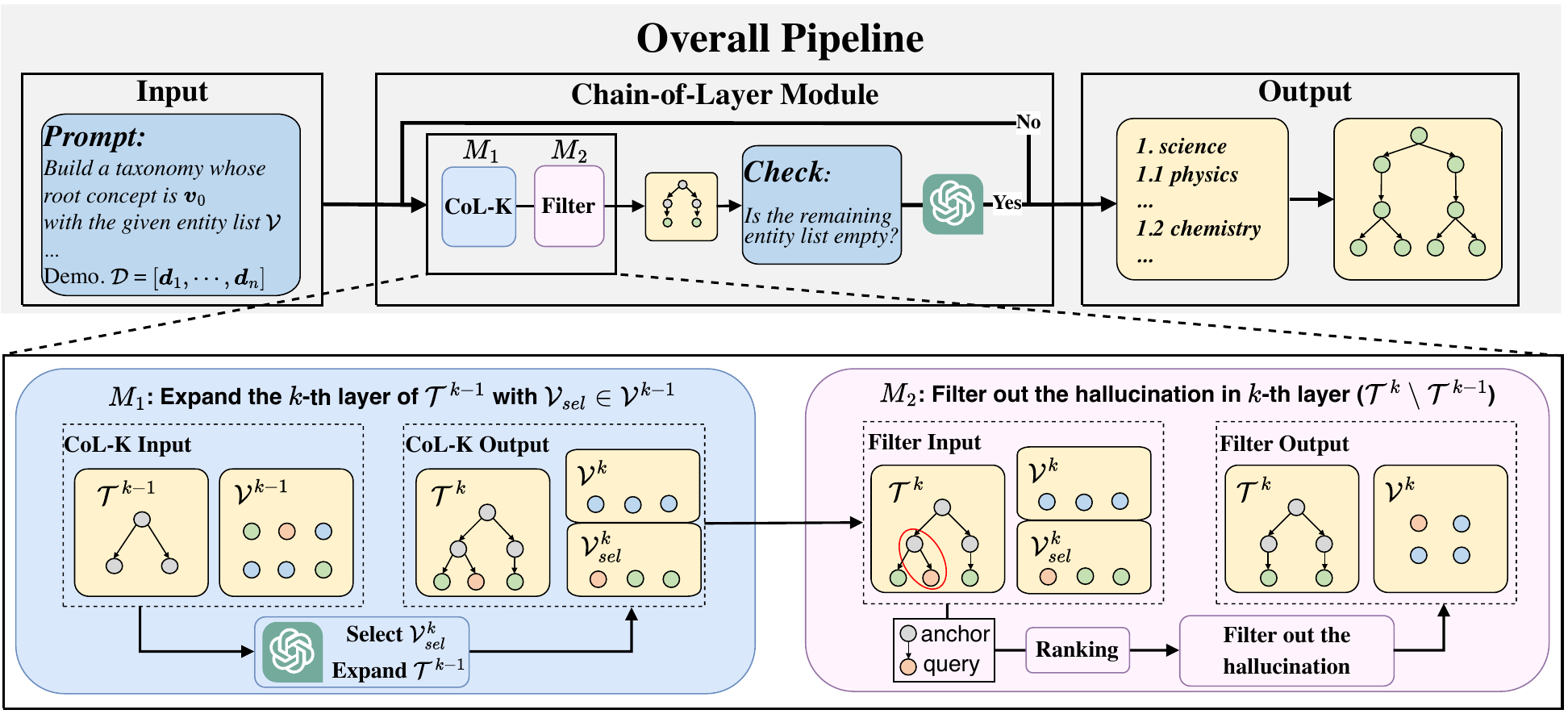}
    \caption{The overview of the framework for \textsc{Chain-of-Layer (CoL)}: Given an entity list $\mathcal{V}$ and a root entity $v_0 \in \mathcal{V}$, \textsc{CoL} systematically organizes the entities in $\mathcal{V}$ into hierarchical groups, incrementally adding them to the taxonomy in a top-down manner at each iteration. In detail, at the $k$-th iteration, \textsc{CoL-K} selects a subset of entities $\mathcal{V}{\text{sel}}$ from the k-level and extends the existing taxonomy $\mathcal{T}^{k-1}$ with these entities. The newly generated parent-child relations ($\mathcal{T}^{k} \setminus \mathcal{T}^{k-1}$) are refined by an Ensemble-based Ranking Filter to reduce the hallucinations into the output taxonomy $\mathcal{T}^{k}$ in $k$-th iteration. The process continues until all entities in $\mathcal{V}$ are integrated into the resulting taxonomy.}
    \label{fig:overview}
    \vspace{-0.1in}
\end{figure*}

Traditional approaches in taxonomy induction follow the discriminative method illustrated in Figure~\ref{fig:motivation_sub1} and aim to identify and structure parent-child relations among entities in a hierarchical manner. Early efforts involve learning these relations by leveraging the semantic connections between entities. The semantics can be represented by lexical patterns~\cite{snow2004learning,kozareva2010semi,panchenko2016taxi,zeng2019faceted}, distributional word embeddings~\cite{fu2014learning,luu2016learning,shwartz2016improving,shang2020taxonomy}, and contextual pre-trained models~\cite{chen2021constructing, jain2022distilling}. Following this, the identified relations are organized into a taxonomic structure using various pruning techniques~\cite{bansal2014structured,velardi2013ontolearn,panchenko2016taxi,mao2018end}. 

Recently, Large Language Models (LLMs) have shown impressive skills in understanding and generating text, enabling them to adapt to a wide range of domains and tasks~\cite{ouyang2022training,achiam2023gpt}. Consequently, many studies have been conducted to leverage the capabilities of LLMs for Information Extraction (IE) tasks using a generative approach~\cite{xu2023large}. Furthermore, increasing the number of parameters of LLMs significantly enhances their ability to generalize, surpassing smaller pre-trained models, and enabling them to deliver outstanding performance in few-shot or zero-shot settings~\cite{kaplan2020scaling}. Figure~\ref{fig:motivation_sub2} illustrates the pipeline depicting how generative methods operate on the taxonomy induction task. 

In the context of taxonomy induction with large language models, TaxonomyGPT~\cite{chen2023prompting} first attempts to prompt LLMs to predict the hierarchical relation among the given concepts. However, TaxonomyGPT shows two major limitations in taxonomy induction. First, it ignores the inherent structure of taxonomies during the generation of new parent-child relations. The reason is that TaxonomyGPT produces parent-child relations among given entities independently, leading to the loss of crucial taxonomic structure information, such as sibling-sibling and ancestor-descendant relations. Consequently, this neglect results in structural inaccuracies in the output taxonomy, including the emergence of multiple root entities and circular relations. Second, as with all the methods based on prompting LLMs, TaxonomyGPT also suffers from the issue of hallucination. For example, even though we have highlighted the requirement in the instruction, LLMs still add entities that are not related to the target taxonomy into the output taxonomy. 

To address the above issues, we first introduce \textsc{HF}, \textbf{H}ierarchical \textbf{F}ormat Taxonomy Induction Instruction to represent taxonomic structures via hierarchical numbering format. For example, as shown in Figure~\ref{fig:motivation_sub2}, science is the sole root entity, so it is indexed as `1. science'. Physics and chemistry, being child entities of science, are thus indexed as `1.1 physics' and `1.2 chemistry'. This format ensures that each entity within the taxonomy possesses a global view of its hierarchical structure, like physics is the sibling entity of chemistry since they share the same hierarchical format (`1.x'). It is important to note that all the methods proposed in this work adhere to \textsc{HF} for representing the taxonomic structure.

Second, to reduce the hallucination generated by the inductive process, we propose the \textsc{Chain-of-Layer (CoL)} unlike prompting LLMs to generate target taxonomy in one iteration, \textsc{CoL} decomposes the taxonomy induction task in a layer-to-layer manner. Specifically, for each iteration, \textsc{CoL} selects a subset of entities from the given entity set and expands the current taxonomy with these selected entities. The key insight of this decomposition is to instruct the LLMs to explicitly anchor each of their reasoning iterations in the taxonomy induction task. Benefits on the iterative setting of \textsc{CoL}, we incorporate an Ensemble-based Ranking Filter at each iteration as a post-processing module to reduce the error propagation from the current iteration to the next iteration. We also develop \textsc{CoL-Zero} to extend \textsc{CoL} to zero-shot settings where annotated taxonomies are unavailable. \textsc{CoL-Zero} uses LLMs to generate taxonomies as demonstrations instead of relying on human-annotated taxonomies.

\begin{figure}[h]
    \centering
    \includegraphics[width=1.0\linewidth]{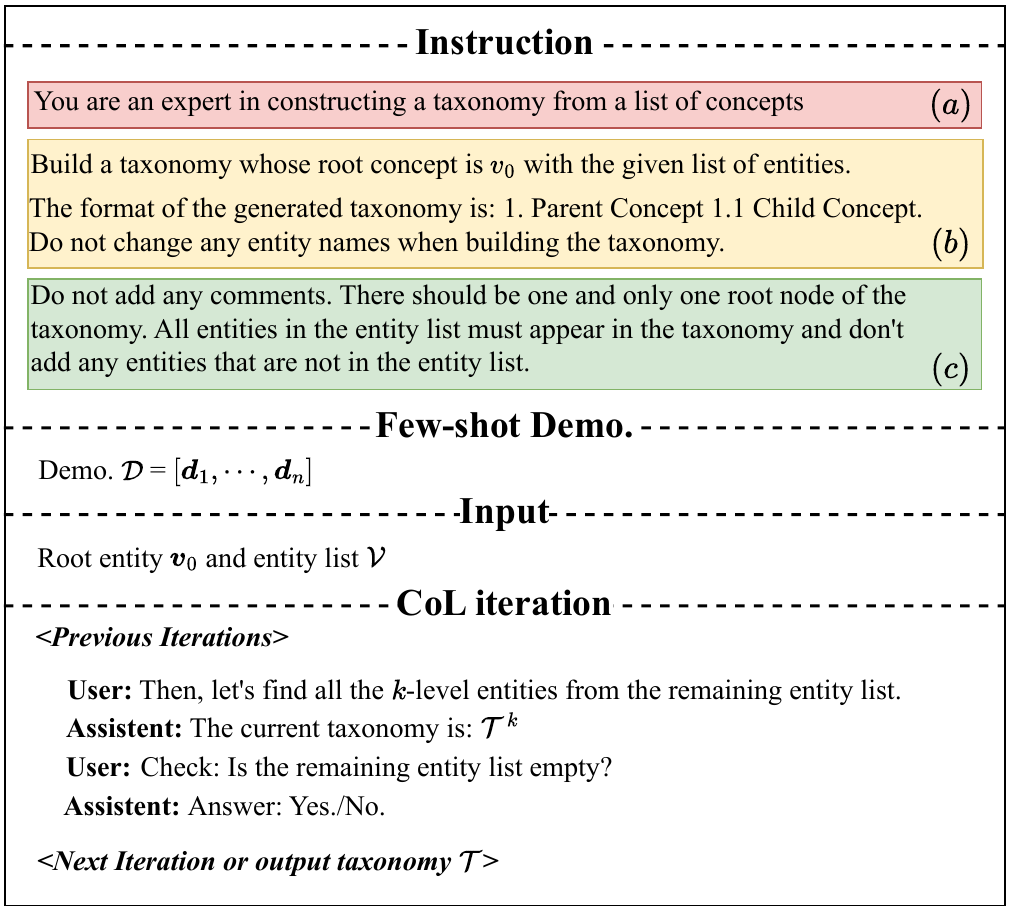}
    \caption{Prompt Overview of \textsc{Chain-of-Layer} Framework}
    \label{fig:overview_prompt}
    \vspace{-0.25in}
\end{figure}

The efficacy of \textsc{HF} and \textsc{CoL} has been validated through extensive experiments on {WordNet} sub-taxonomies and three large-scale, real-world taxonomies. The results demonstrate that both \textsc{HF} and \textsc{CoL} outperform all baseline methods across multiple evaluation metrics. We also explore the performance of \textsc{CoL-Zero} on the benchmarks mentioned above. Some interesting observations of \textsc{CoL-Zero} are presented in this work.


In summary, this study makes the following contributions:
\begin{compactitem}
    \item We introduce \textsc{HF}, the Hierarchical Format Taxonomy Induction Instruction, to utilize the hierarchical structure of the entities to increase the quality of the inducted taxonomy. 
    \item We introduce \textsc{Chain-of-Layer (CoL)}, an iterative taxonomy induction framework that incorporates the Ensemble-based Ranking Filter for reducing the hallucinations in the output taxonomies generated by LLMs.
    \item Extensive experiments demonstrate that \textsc{HF} and \textsc{CoL} significantly improve the performance of taxonomy induction tasks on four datasets from various domains.
\end{compactitem}

\indent \textbf{Scope and Limitation.} This study represents an initial effort to utilize LLMs for taxonomy induction. Our main focus is to identify an effective in-context learning framework to harness the capabilities of LLMs. We are aware that the performance of our proposed approach on large-scale taxonomies is constrained by the limitations in instruction-following capabilities and the context window size of LLMs. However, how to facilitate the ability of LLMs to handle extremely long prompts is beyond the scope of this paper. We hope this work will inspire future research in this area.


\label{sec:intro}

\section{Problem Definition}
We define a taxonomy, denoted as $\mathcal{T} = (\mathcal{V}, \mathcal{E})$, as a directed acyclic graph composed of two components: a vertex set $\mathcal{V}$ and an edge set $\mathcal{E}$. In the task of taxonomy induction, the model is provided with a set of conceptual entities, represented by $\mathcal{V}$, where each entity can be either a single word or a short phrase. The objective is to construct the taxonomy $\mathcal{T}$ based on these given entities.


\section{Methodology}

In this section, we provide a comprehensive overview of our proposed \textsc{Chain-of-Layer (CoL)} framework designed for addressing the taxonomy induction task. Specifically, \textsc{CoL} dissects the taxonomy induction task through a layer-to-layer approach. As shown in Figure \ref{fig:overview_prompt}, Our \textsc{CoL} framework consists of four parts: instruction (\textsc{HF}, \textbf{H}ierarchical \textbf{F}ormat Taxonomy Induction Instruction), few-shot demonstration, input, and \textsc{CoL} iteration. In the instruction part (Sec.~\ref{sec:task-description}), we configure the system message of the LLM, specify the objectives of the task and the expected output format, and establish a series of rules that help the model understand and accurately complete the task. In Sec.~\ref{sec:demo-design}, we describe and formalize the process of inducting our demonstrations. In Sec.~\ref{sec:infer}, we introduce the iterative process of \textsc{CoL} and the Ensemble-based Ranking Filters tailored to mitigate hallucinations that may arise during the process. Finally, in Sec.~\ref{sec:col-zero}, we extend our  \textsc{CoL} to the zero-shot setting. The details of each module in \textsc{CoL} are presented in Figure~\ref{fig:overview}.

\subsection{Hierarchical Format Taxonomy Induction Instruction (HF)}
\label{sec:col_design}
\label{sec:task-description}
To enable LLMs to more effectively and accurately complete the taxonomy induction task, we propose \textsc{HF}, the \textbf{H}ierarchical \textbf{F}ormat Taxonomy Induction Instruction. As shown in Figure \ref{fig:overview_prompt}, the instruction specifies the objectives of the taxonomy induction task, which can be decomposed into three components. In component (\textit{a}), LLMs are instructed to utilize the domain expertise to generate the desired output. Component (\textit{b}) provides instructions for the output format, which is expected to adhere to a hierarchical numbering format. This format ensures that each entity within the generated taxonomy possesses a comprehensive understanding of its hierarchical structure. Finally, component (\textit{c}) highlights a set of fundamental rules $\mathcal{R}$ about the taxonomy induction task. These rules include: 1. Do not use entities not covered in the given entity set and ensure that all entities listed in the given entity list are present in the taxonomy ($\boldsymbol{r}_1$); 2. Maintain a single root entity within the taxonomy ($\boldsymbol{r}_2$); 3. Refrain from adding comments ($\boldsymbol{r}_3$).

\subsection{Few-shot Demonstration Construction}
\label{sec:demo-design}

To enable the model to better follow our instructions to complete the task, we propose a method for constructing demonstrations for \textsc{CoL} inference. For each demonstration $d_i$, we decompose each taxonomy $\mathcal{T}_{i}$ in hierarchical order and simulate the process of inducting the entire taxonomy from top to bottom. At the end of each level of induction, we prompt LLM whether the current taxonomy has included all entities from the given entity set. If a negative response is received, we will continue to expand the current taxonomy layer downward until it encompasses all entities from the given entity set. The demonstration $d_i$ employed for expanding the $k$-th layer of the demo taxonomy $\mathcal{T}^{k-1}_{\boldsymbol{d}_i}$ are presented as follows:
\begin{quote}
    \textit{<messages of previous iteration>} \\
    \textbf{Assistant:} \textit{The current taxonomy is:} $\mathcal{T}^{k-1}_{\boldsymbol{d}_i}$\\
    \textbf{User:} \textit{Check: Is the remaining entity list empty?}\\
    \textbf{Assistant:} \textit{Answer: No.} \\
    \textbf{User:} \textit{Then, let's find all the $k$-th level entities from the remaining entity list.}\\
    \textbf{Assistant:} \textit{The current taxonomy is:} $\mathcal{T}^{k}_{\boldsymbol{d}_i} \gets \mathcal{T}^{k-1}_{\boldsymbol{d}_i} \bigcup \mathcal{V}^{k}_{sel_i}$ \\
    \textit{<beginning of next iteration>} \\
\end{quote}
We use the first five sub-taxonomies from WordNet's training set as demonstrations $\mathcal{D}^{(t)}= [\boldsymbol{d}_1^{(t)},\cdots,\boldsymbol{d}_5^{(t)}]$ in this case for fair comparision. 

\begin{figure}[t]
    \centering
    \includegraphics[width=1.0\linewidth]{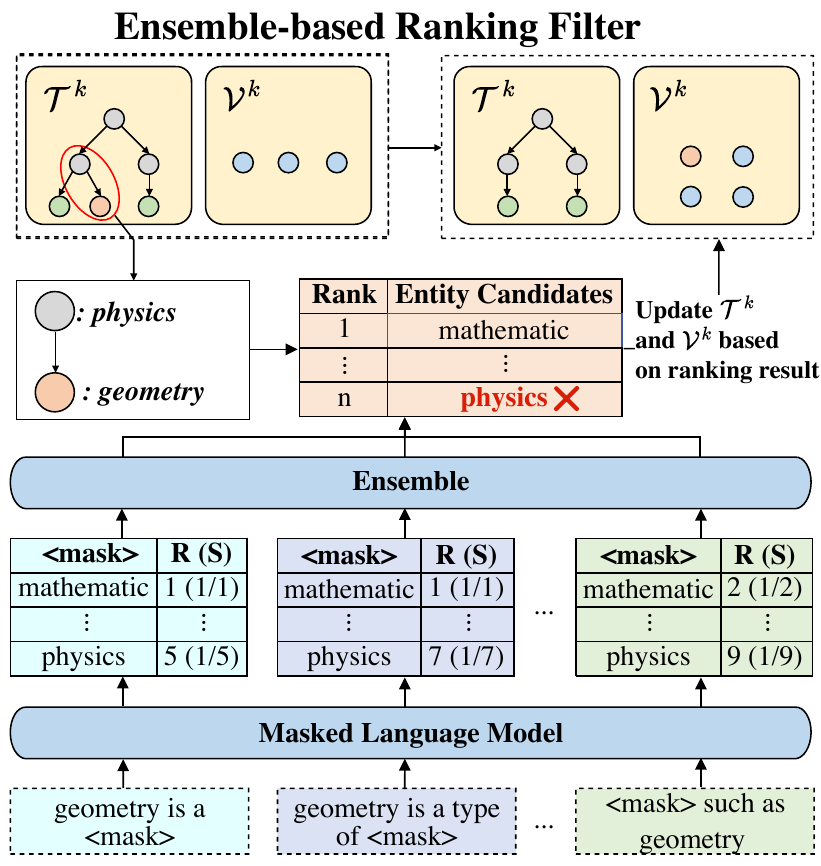}
    \caption{The details of the Ensemble-based Ranking Filter.}
    \label{fig:overvier_filter}
    \vspace{-0.2in}
\end{figure}

\subsection{Inference via \textsc{Chain-of-Layer}}
\label{sec:infer}
\subsubsection{Ensemble-based Ranking Filter}
\label{sec:filter}
It is well-known that large language models are greatly affected by hallucinations during the process of generating target texts, resulting in content that is significantly different from the target~\cite{ji2023survey, zhang2023siren}. In the taxonomy induction task, we mainly observe two categories of LLM hallucinations: (1) The large language models do not strictly use the entities in the given entity set but instead include non-target entities in the output taxonomy; (2) The large language models introduce incorrect parent-child relations into the output taxonomy. 

To alleviate these issues, we propose a filter module in the \textsc{CoL} framework. Specifically, in the process of inducting each layer of the taxonomy, the filter removes incorrect parent-child relations in each iteration of the model's output, preventing the error caused by hallucination propagating to the next iteration.

Our filter design is based on an ensemble mechanism. We propose a set of templates $\mathcal{M}$ and used a pre-trained mask language model to rank the generated parent-child relations in each iteration for all templates $\boldsymbol{m} \in \mathcal{M}$. 
We present the $\mathcal{M}$ as follows:
\begin{quote}
    \textit{\texttt{<query>} is a/an \texttt{<anchor>}} \\
    \textit{\texttt{<query>} is a kind of \texttt{<anchor>}} \\
    \textit{\texttt{<query>} is a type of \texttt{<anchor>}} \\
    \textit{\texttt{<query>} is an example of \texttt{<anchor>}} \\
    \textit{\texttt{<anchor>} such as \texttt{<query>}} \\
    \textit{A/An \texttt{<anchor>} such as \texttt{<query>}} \\
\end{quote}
For each entity $\boldsymbol{q}$ in the entity list, we compute the probability of tokens at the \texttt{<anchor>} position by placing $\boldsymbol{q}$ in the \texttt{<query>} position (as a child entity) of template $\boldsymbol{m}$ and positioning one of the remaining entities $\boldsymbol{a}$ at the \texttt{<anchor>} (as a parent entity). Then we sort the token probabilities to determine the similarity ranking. Subsequently, inference, represented by $\mathrm{Sim}(\boldsymbol{q}, \boldsymbol{a} | \boldsymbol{m})$, is computed utilizing the reciprocal of this similarity ranking.
Finally, we ensemble the scoring results of each template to obtain the final filter score. The formula of the similarity score is: 
\begin{align}
    \textit{score}(\boldsymbol{q} | \boldsymbol{a}, \mathcal{M},\mathcal{V}) = \frac{1}{\left | \mathcal{M}  \right | } \sum_{\boldsymbol{m} \in \mathcal{M}}^{}\mathrm{Sim}(\boldsymbol{q}, \boldsymbol{a} | \boldsymbol{m}) 
\label{eq:scoring}
\end{align}

For each query entity, we retain only the top ten parent candidates. If the parent-child relations output by the LLM are not within this range, these relations will be filtered out. In this paper, we use a pre-trained masked language model specialized for the scientific domain and tasks called SciBERT~\cite{beltagy2019scibert} to ensure that the pre-trained models contain sufficient domain knowledge to complete the ranking process. 

\subsubsection{Iterative Inference}
\label{sec:infer_col}
After providing the instructions and constructing the few-shot demonstrations, we introduce the interactive inference process of our \textsc{CoL} framework with the Ensemble-based Ranking Filter. We provide the input entity candidates set $\mathcal{V}$ and the initial taxonomy $\mathcal{T}^{0}$ only including root entity $\boldsymbol{v}_0$ to the LLM and expect the LLM to generate the output taxonomy $\mathcal{T}$ according to the rules $\mathcal{R}$ and the defined format of the demonstrations $\mathcal{D}$ in section~\ref{sec:demo-design}. The whole inference process in \textsc{CoL} is donated as:
\begin{align}
    \mathcal{T}  =  \mathrm{CoL} (\mathcal{V}, \mathcal{T}^{0}, \mathcal{D}, \mathcal{R})
\end{align}

The inference starts with $k = 0$ and define $\mathcal{T}^{0} = \boldsymbol{v}_0$, $\mathcal{V}^{0}_{sel} = [\boldsymbol{v}_0]$. And in the $k$-th iteration, we first prompt the LLM to generate the $k$-th layer of the taxonomy, and update taxonomy $\mathcal{T}^{k-1}$ to $\mathcal{T}^{k}$, and remaining entity list $\mathcal{V}^{k-1}$ to $\mathcal{V}^{k}$. The $k$-th inference process in \textsc{CoL} is donated as:
\begin{align}
    \mathcal{T}^{k}, \mathcal{V}^{k}_{sel} &=  \mathrm{CoL-K} (\mathcal{V}^{k-1}, \mathcal{T}^{k-1}, \mathcal{D}, \mathcal{R}) \\
    \mathcal{V}^{k} &= \mathcal{V}^{k-1} \setminus \mathcal{V}^{k}_{sel}
\end{align}

To alleviate the impact of the model's hallucinations on the quality of the output taxonomy, we employ the Ensemble-based Ranking Filter to filter out the hallucinations in the generated parent-child relations which are in $(\mathcal{T}^k \setminus \mathcal{T}^{k-1})$ at $k$-th iteration. Then we update the output taxonomy $\mathcal{T}^{k}$ and the remaining entity list $\mathcal{V}^{k}$ at $k$-th iteration. The processing of the Ensemble-based Ranking Filter is donated as:
\begin{align}
    \mathcal{T}^{k}, \mathcal{V}^{k}  \leftarrow \mathrm{EnsembleFilter} (\mathcal{T}^{k}, \mathcal{V}^{k}, \mathcal{V}^{k}_{sel})
\end{align}

At the end of each iteration, we prompt the model to check if the remaining entity list $\mathcal{V}^{k}$ is empty or not. If we receive a positive response, we then output the current taxonomy $\mathcal{T}^{k}$ as the final result $\mathcal{T}$. Otherwise, we proceed to the next iteration.

\subsection{Demonstrations Generation via LLMs}
\label{sec:col-zero}

While \textsc{CoL} is designed to induct taxonomy from given entity sets via the few-shot learning setting. In domains that lack well-inducted taxonomies, we propose a zero-shot \textsc{CoL} alternative \textsc{CoL-Zero}. The idea of \textsc{CoL-Zero} is utilizing LLMs to generate taxonomies instead of utilizing taxonomies annotated by human experts acts as demonstrations. 

The details of \textsc{CoL-Zero} are as follows. We start with the root entity $v_0$ of the target taxonomy $\mathcal{T}$. We follow the instruction mentioned in section~\ref{sec:task-description} to prompt LLM directly to generate the taxonomies $\mathcal{T}_{\boldsymbol{d}^{(g)}_{i}}$, used to construct demonstration $\boldsymbol{d}_i^{(g)}$ following the process in section \ref{sec:demo-design}. Thus we have $\mathcal{D}^{(g)} = [\boldsymbol{d}_1^{(g)},\cdots,\boldsymbol{d}_5^{(g)}]$. Different from \textsc{CoL}, \textsc{CoL-Zero} remove the restriction of only using the entities that are covered in the given entity set ($r_1$) to free form $\mathcal{R}^{\prime} = [r_2, r_3]$.
\begin{align}
    \mathcal{T}_{\boldsymbol{d}^{(g)}_{i}} &= \mathrm{CoL-demo-generation} (\mathcal{T}^{0}, \mathcal{R}^{\prime}) \\
    \mathcal{T} &= \mathrm{CoL} (\mathcal{V}, \mathcal{T}^{0}, \mathcal{D}^{(g)}, \mathcal{R}^{\prime})
\label{eq:zero-shot-infer}
\end{align}

\label{subsec:prompt}

\section{Experiments}

Our proposed \textsc{HF} and \textsc{CoL} are evaluated on four benchmarks. The experiments aim to address three research questions (RQs):
\begin{compactitem}
    \item \textbf{RQ1:} How does the performance of the proposed framework compare to state-of-the-art baselines in taxonomy induction?
    \item \textbf{RQ2:} How does the proposed framework perform on scalability and domain generalization?
    \item \textbf{RQ3:} Which components within the proposed framework most significantly impact the effectiveness of taxonomy induction tasks? How can the hyperparameters for these components be determined?
\end{compactitem}

\begin{table}[t]
    \centering
    \Scale[1.0]{\begin{tabular}{lccc}
    \toprule
    & \#Concepts & \#Edges & Depth \\
    \midrule
     WordNet & 20.5 / 20.5  & 19.5 / 19.5 & 3.0 / 3.0 \\ 
    Wiki & 102.6 / 252.0 & 101.8 / 255.0 & 2.2 / 3.0 \\ 
    DBLP & 90.8 / 176.0 & 89.8 /175.0 & 2.8 / 4.0 \\ 
    SemEval-Sci & 114.0 / 429.0 & 113.0 / 451.0  & 7.2 / 8.0 \\ \bottomrule
    \end{tabular}}
    \caption{Statistics of four taxonomy datasets. Each cell is presented as */*, indicating the average for sampled sub-taxonomies and the entire taxonomy, respectively.}
    \label{tab:datasets}
    \vspace{-0.35in}
\end{table}

\begin{table}[t]
    \centering
    \resizebox{1\linewidth}{!}{
    \begin{tabular}{lcccccc}
         \toprule[1.5pt]
         \multirow{2}{*}{\textbf{Model}} & \multicolumn{6}{c}{\textbf{WordNet}}\\
         \cmidrule(lr){2-7}
         & $\textbf{P}_a$ & $\textbf{R}_a$ & $\textbf{F1}_a$ & $\textbf{P}_e$ & $\textbf{R}_e$ & $\textbf{F1}_e$\\
         \midrule[0.75pt]
         \multirow{2}{*}{} & \multicolumn{3}{c}{\textit{Supervised Fine-tuning}} & \multicolumn{2}{c}{}\\
         \textsc{Graph2Taxo}~\cite{shang2020taxonomy} & 79.20 & 47.80 & 59.60 & 75.60 & 37.00 & 49.70  \\
          \textsc{CTP}~\cite{chen2021constructing} & 69.30 & 66.20 & 66.70 & 53.30 & 49.80 & 51.50 \\
          \textsc{CTP-Llama-2-7B}~\cite{chen2021constructing} & 73.48 &  70.02 & 71.71 & 55.42 & 51.98 & 53.64 \\
          
        \midrule[0.75pt]

          \multirow{2}{*}{} & \multicolumn{3}{c}{\textit{Zero-shot Setting}} & \multicolumn{2}{c}{} \\
         \textsc{RestrictMLM}~\cite{jain2022distilling} & 23.23 & 25.69 & 24.09  & 24.17 & 25.65 & 24.89 \\
          \textsc{LMScorer}~\cite{jain2022distilling} & 37.50 & 47.64 & 41.59 & 36.27 & 38.48 & 37.34  \\
          \hdashline
          \multirow{2}{*}{} & \multicolumn{3}{c}{\textit{Ours}} & \multicolumn{2}{c}{} \\
          \textsc{HF (GPT-4)} & 81.13 & \textbf{78.35} & \textbf{78.37} & \underline{53.27} & \underline{54.63} & \underline{53.87} \\
          \textsc{HF (GPT-3.5)} & 85.92 & 61.75 & 69.62 &  45.91 & 43.57 & 44.15 \\
          \textsc{CoL-zero (GPT-4)} & \textbf{89.71} & \underline{71.39} & \underline{78.31} & \textbf{58.93} & \textbf{55.18} & \textbf{56.41} \\
          \textsc{CoL-zero (GPT-3.5)} & \underline{86.92} & 60.06 & 69.61 & 48.78 & 42.04 & 44.50 \\
          \midrule[0.75pt]
          
          \multirow{2}{*}{} & \multicolumn{3}{c}{\textit{5-shot Setting}} & \multicolumn{2}{c}{} \\
          \textsc{TaxonomyGPT (GPT-4)}~\cite{chen2023prompting} & 53.09 & 31.84 & 39.07 & 39.59 & 36.84 & 38.01 \\
          \textsc{TaxonomyGPT (GPT-3.5)}~\cite{chen2023prompting} &  62.97 & 41.77 & 48.95 & 49.20 & 43.85 & 46.24\\
        \hdashline
          \multirow{2}{*}{} & \multicolumn{3}{c}{\textit{Ours}} & \multicolumn{2}{c}{} \\
          \textsc{HF (GPT-4)} &  85.33 & \textbf{79.30} & \textbf{81.58} & \underline{58.96} & \textbf{59.22} & \textbf{59.08} \\
          \textsc{HF (GPT-3.5)} & 80.48 & 72.59 & 75.37  & 49.95 & 49.26 & 49.46 \\              
          \textsc{CoL (GPT-4)} & \textbf{90.60} & \underline{73.07} & \underline{79.62}  & \textbf{59.57} & \underline{57.10} & \underline{57.73} \\
          \textsc{CoL (GPT-3.5)} &  \underline{85.69} & 60.16 & 69.39 & 47.90 & 41.92 & 44.26 \\
        \bottomrule[1.5pt]
    \end{tabular}
    }
    \caption{Performance comparison across WordNet sub-taxonomies in three different settings: Bold indicates the highest performance within each setting, while underlined denotes the second best performance within each setting.}
    \label{tab:Results_Wordnet}
    \vspace{-0.35in}
\end{table}

\subsection{Experimental Setting}
\subsubsection{Datasets}
We conducted our experiments using WordNet sub-taxonomies created by~\cite{bansal2014structured}. This dataset comprises 761 non-overlapping taxonomies, each with 11 to 50 entities and the depth of each sub-taxonomy is 4. It means there are 4 entities along the longest path from the root entity to any leaf entity. The WordNet is divided into training (533), development (114), and test (114) sets.

Furthermore, we evaluate our framework using three large-scale real-world taxonomies: (1) \textbf{DBLP} is constructed from 156,000 computer science paper abstracts; (2) \textbf{Wiki} is derived from a subset of English Wikipedia pages; (3) \textbf{SemEval2016-Sci} is derived from the shared task of taxonomy induction in SemEval2016. For \textbf{DBLP} and \textbf{Wiki}, we uses annotation results from~\cite{shen2017setexpan}.

Due to the sequence length limitation of LLMs, we conducted five separate samplings for these three large-scale taxonomies, ensuring that the size of each sampled sub-taxonomy ranged from 80 to 120 entities. The experimental results are averaged over these five samplings. The dataset statistics are presented in Table~\ref{tab:datasets}.

\subsubsection{Baseline Methods} 

We compare the proposed framework with the following supervised fine-tuning baseline methods:
\begin{compactitem}
    \item \textbf{Graph2Taxo}~\cite{shang2020taxonomy}: leverages cross-domain graph structures and adopts constraint-based Directed Acyclic Graph (DAG) learning for taxonomy induction.
    \item \textbf{CTP}~\cite{chen2021constructing}: fine-tunes RoBERTa model to predict parent-child pair likelihoods and integrates these into a graph using a maximum spanning tree algorithm for precise taxonomy induction. Additionally, we present results using a Llama-2-7B model as the backbone for CTP.
\end{compactitem}

We compare the following unsupervised and in-context learning baseline methods:
\begin{compactitem}
    \item \textbf{RestrictMLM}~\cite{jain2022distilling}: utilizes a cloze statement, or 'fill-in-the-blank', method to extract 'is-a' relational knowledge from BERT. However, this approach is limited to single-gram entities due to the constraints of the schema.
    \item \textbf{LMScore}~\cite{jain2022distilling}:  treats taxonomy induction as a sentence scoring task using GPT-2. It assesses the natural fluency of sentences that elicit parent-child relations.
    \item \textbf{TaxonomyGPT}~\cite{chen2023prompting}: approaches taxonomy induction as a conditional text generation challenge. It represents the output taxonomy as a collection of sentences, each describing a parent-child relation within the output taxonomy. 
\end{compactitem}

For our proposed framework, we conduct experiments with \textsc{GPT-3.5-turbo-16k} and \textsc{GPT-4-1106-preview}\footnote{To facilitate the following discussion, we abbreviated them as GPT-3.5 and GPT-4.}. For \textsc{HF}, we directly prompt the LLMs using the \textsc{HF} instruct describe in Section \ref{sec:task-description}.

\begin{table*}[ht]
    \centering
    \resizebox{1\linewidth}{!}{
    \begin{tabular}{lcccccccccccccccccc}
         \toprule[1.5pt]
         \multirow{2}{*}{\textbf{Model}} & \multicolumn{6}{c}{\textbf{Wiki}} & 
         \multicolumn{6}{c}{\textbf{DBLP}} & \multicolumn{6}{c}{\textbf{SemEval-Sci}}\\
         \cmidrule(lr){2-7}
         \cmidrule(lr){8-13}
         \cmidrule(lr){14-19}
         & $\textbf{P}_a$ & $\textbf{R}_a$ & $\textbf{F1}_a$ & $\textbf{P}_e$ & $\textbf{R}_e$ & $\textbf{F1}_e$
         & $\textbf{P}_a$ & $\textbf{R}_a$ & $\textbf{F1}_a$ & $\textbf{P}_e$ & $\textbf{R}_e$ & $\textbf{F1}_e$
         & $\textbf{P}_a$ & $\textbf{R}_a$ & $\textbf{F1}_a$ & $\textbf{P}_e$ & $\textbf{R}_e$ & $\textbf{F1}_e$\\
         
         \midrule[0.75pt]
         \multirow{2}{*}{} & \multicolumn{6}{c}{} & 
         \multicolumn{6}{c}{\textit{Supervised Fine-tuning}} & \multicolumn{6}{c}{}\\
        \textsc{Graph2Taxo}~\cite{shang2020taxonomy} & 43.02 & 36.50 & 39.49 & 39.28 & 34.12 & 36.52 & 47.85 & 30.23 & 37.05 & 46.63 & 28.49 & 35.37 & 82.45 & 36.15 & 50.27 & 79.37 & 34.52  & 46.87  \\
        \textsc{CTP}~\cite{chen2021constructing} & 50.94 & 47.15 & 48.97  & 46.56 & 42.53 & 44.45 &  45.62 & 41.39 & 43.40 & 38.21 & 33.73 & 35.83  & 52.41 & 33.88  & 41.16 & 31.18 & 29.42 & 30.27 \\ 
        \textsc{CTP-Llama-2-7B}~\cite{chen2021constructing} & 67.74 & 64.16 & 65.78 & 63.64 & 60.07 & 61.80 & 48.73 & 39.88 & 43.86 & 44.39 & 35.81 & 39.64 & 61.98 & 54.09  & 57.77 & 48.33 & 41.92 & 44.90 \\ 
        \midrule[0.75pt]
         \multirow{2}{*}{} & \multicolumn{6}{c}{} & 
         \multicolumn{6}{c}{\textit{Zero-shot Setting}} & \multicolumn{6}{c}{}\\        \textsc{RestrictMLM}~\cite{jain2022distilling} & 49.88 & 54.08 & 51.85 & 30.01 & 30.21 & 30.11 & - & - & - & - & - & - &  63.33 & 47.85 & 54.44 & 45.79 & 46.19 & 45.99 \\         
        \textsc{LMScorer}~\cite{jain2022distilling} & 18.77 & 25.94 & 21.74 & 19.78 & 19.95 & 19.86 & 17.14 & 21.54 & 19.04 & 25.84 & 26.12 & 25.98 & 48.80 & 33.24 & 39.51 & 42.20 & 42.58 & 42.39 \\
        \hdashline
        \multirow{2}{*}{} & \multicolumn{6}{c}{} &  \multicolumn{6}{c}{\textit{Ours}} & \multicolumn{6}{c}{}\\  
        \textsc{HF (GPT-4)} & \underline{92.96} & \textbf{94.48} & \textbf{93.68} & 91.55 & \textbf{91.31} & \textbf{91.41} & 52.70 & \textbf{64.69} & \textbf{57.65}  & 30.76 & 29.58 & 29.91 & 78.56 & \textbf{54.68} & \textbf{64.02} & 45.12 & \textbf{46.64} & \underline{45.85} \\
        \textsc{HF (GPT-3.5)} & 75.85 & 71.55 & 73.36 & 71.67 & 73.63 & 72.03 & 50.20 & 48.28 & 48.76 & 27.66 & 26.51 & 26.98 &  70.25 & 40.91 & 51.17 & 28.69 & 28.11 & 28.29 \\       
        \textsc{CoL-zero (GPT-4)} & \textbf{100.00} & \underline{84.58} & \underline{91.12} & \textbf{99.70} & \underline{84.77} & \underline{91.15} & \textbf{80.88} & \underline{54.25} & \underline{57.21} & \underline{40.15} & \textbf{35.88} & \underline{37.81} & \textbf{94.99} & \underline{45.83} & \underline{61.66} & \textbf{62.33} & \underline{45.55} & \textbf{52.44} \\
        \textsc{CoL-zero (GPT-3.5)} & 99.72 & 57.92 & 72.65 &  \underline{99.17} & 58.23 & 72.76 & \underline{76.78} & 38.39 & 49.36 & \textbf{53.72} & \underline{30.61} & \textbf{38.02} & \underline{93.12} & 22.43 & 35.59 & \underline{56.52} & 22.13 & 31.54 \\
          
        \midrule[0.75pt]
         \multirow{2}{*}{} & \multicolumn{6}{c}{} & 
         \multicolumn{6}{c}{\textit{5-shot Setting}} & \multicolumn{6}{c}{}\\           
         \textsc{TaxonomyGPT}~\cite{chen2023prompting} & 69.26 & 63.48 & 65.19 & 89.55 & 86.71 & 87.98 &  28.98 & 14.40 & 17.15 & 34.27 & 22.17 & 25.97 &  53.09 & 31.84 & 39.07 & 39.59 & 36.84 & 38.01 \\ 
        \hdashline
         \multirow{2}{*}{} & \multicolumn{6}{c}{} & \multicolumn{6}{c}{\textit{Ours}} & \multicolumn{6}{c}{}\\  
         
        \textsc{HF-Zero (GPT-4)} & 96.33 & \underline{95.18} & \underline{95.75} & 93.08 & \underline{91.72} & \underline{92.39} & 59.76 & \textbf{74.37} & \underline{65.83} & 38.42 & \underline{40.20} & \underline{39.28} & 75.28 & \textbf{59.32} & \underline{62.63} & 43.64 & \textbf{49.29} & \underline{45.24}\\
        \textsc{HF-Zero (GPT-3.5)} & 88.88 & 80.36 & 84.38 & 83.67 & 74.92 & 78.98 & 62.42 & 53.76 & 57.53 & 32.68 & 28.59 & 30.38 & 57.00 & 36.89 & 44.38  & 29.51 & 29.88 & 29.35\\
        
        \textsc{CoL (GPT-4)}  & \textbf{99.17}& \textbf{95.99} & \textbf{97.54} & \textbf{97.92} & \textbf{94.99} & \textbf{96.43} & \textbf{79.95} & \underline{63.06} & \textbf{68.82}  & \underline{55.07} & \textbf{44.27} & \textbf{47.96} & \underline{91.23} & \underline{48.16} & \textbf{62.69} & \underline{59.60} & \underline{46.03} & \textbf{51.59}  \\
        \textsc{CoL (GPT-3.5)} & \underline{99.00} & 73.25 & 83.73  & \underline{97.54} & 71.99 & 82.41 & \underline{79.74} & 42.21 & 54.76 & \textbf{55.35} & 28.70 & 37.66 & \textbf{95.75} & 26.66 & 41.35 & \textbf{59.73} & 26.05 & 35.99 \\
        \bottomrule[1.5pt]
    \end{tabular}
    }
    \caption{Performance on taxonomy induction on three large scale taxonomies: Bold for the highest among all. Underlined for the second-best performance. Due to the scalability challenges discussed in Section~\ref{sec:exp_scalability}, each method was applied to five sub-taxonomies derived from the original, with results averaged. The \textsc{RestrictMLM} results for DBLP are unavailable since it only handles single-gram entities using a 'fill-in-the-blanks' schema. Due to \textsc{GPT-4} not following the instructions of the \textsc{TaxonomyGPT}'s prompt, only the results from \textsc{TaxonomyGPT (GPT-3.5)} were retained.}
    \label{tab:Results_Large-scale}
    \vspace{-0.3in}
\end{table*}

\subsubsection{Evaluation Metrics}
This section outlines the metrics for evaluating our taxonomy prediction models: Ancestor-F1 and Edge-F1.

\textbf{Ancestor-F1:} This metric assesses ancestor-descendant relations in predicted and ground truth taxonomies.
  \begin{align*}
    P_a &= \frac{\left | \text{is-ancestor}_\text{pred} \bigcap \text{is-ancestor}_\text{gold}  \right | }{\left | \text{is-ancestor}_\text{pred} \right |} \\
    R_a &= \frac{\left | \text{is-ancestor}_\text{pred} \bigcap \text{is-ancestor}_\text{gold}  \right | }{\left | \text{is-ancestor}_\text{gold} \right |} \\
    F1_a &= \frac{2P_a * R_a}{P_a + R_a}
\end{align*}
where $P_a$, $R_a$ and $F1_a$ donate the ancestor precision, recall, and F1-score, respectively. 

\textbf{Edge-F1:} This metric, stricter than Ancestor-F1, compares predicted edges directly with gold standard edges. Edge-based metrics are denoted as $P_e$, $R_e$, and $F1_e$, respectively. 

\subsection{Results on the WordNet (RQ1)}

In our experiments, we compare the performance of our \textsc{HF} and \textsc{CoL} to three major settings baseline methods (supervised fine-tuning, zero-shot setting, and 5-shot setting) on medium-sized WordNet. As the experimental results are shown in Table~\ref{tab:Results_Wordnet}, we have four major observations as follows:

Firstly, \textsc{HF} (GPT-4) and \textsc{CoL} (GPT-4) variants consistently achieved the highest F1 scores, validating the effectiveness of GPT-4 models in taxonomy induction. They significantly outperformed the \textsc{LMScorer} baseline, highlighting the superior text understanding and generation capabilities of LLMs.


Second, despite using powerful LLMs like GPT-4, TaxonomyGPT performed worse than methods such as CTP, which rely on fine-tuning BERT/Llama-2-7B models, across all six metrics. This suggests that LLMs are sensitive to output format requirements. TaxonomyGPT's approach of representing parent-child relationships as independent sentences loses structural coherence. In contrast, \textsc{HF} and \textsc{CoL} use a hierarchical number format to encode positional information, improving performance.

Third, Graph2Taxo achieved the highest precision across all settings, leveraging lexical patterns as direct input features. However, its lower recall indicates a trade-off, suggesting it may not fully capture all taxonomic relations.

Last, comparing \textsc{HF} and \textsc{CoL}, we have the following observations: (1) \textsc{CoL-Zero} (GPT-4) outperforms \textsc{HF-Zero} (GPT-4) with a 9.6\%, 1.0\%, and 4.5\% increase in $P_e$, $R_e$, and $F1_e$. This result demonstrates that \textsc{CoL} is better suited for medium-sized taxonomy induction tasks under the zero-shot setting. (2) Under the 5-shot setting, \textsc{HF} (GPT-4) shows a 2.3\% lead in $F1_e$ compared to \textsc{CoL} (GPT-4), indicating that direct prompting with \textsc{HF} achieves state-of-the-art results when in-domain examples are provided.

\subsection{Results on the Three Large-Scale Taxonomies (RQ1 and RQ2)}



In this section, we present the experiment results on three large-scale taxonomies: Wiki, DBLP, and SemEval-Sci, as shown in Table~\ref{tab:Results_Large-scale}. This experiment tests domain generalization ability, with all supervised fine-tuning trained on WordNet and then tested directly on these taxonomies. Under the 5-shot setting, we use the first five sub-taxonomies of the WordNet training set for a fair comparison. Our observations are as follows:

First, in the supervised fine-tuning setting, models such as \textsc{CTP} and \textsc{Graph2Taxo} provide a foundation for understanding taxonomy induction's intricacies. However, the proposed \textsc{CoL} (GPT-4) shows the best performance in $F1_e$ across all three taxonomies. Compared to the best-performing SFT model, \textsc{CoL} (GPT-4) has increased $F1_e$ by  56.03\%, 20.99\%,  and 10.07\% on Wiki, DBLP and SemEval-Sci, respectively. Compared to \textsc{HF}, \textsc{CoL} (GPT-4) has increased $F1_e$ by  4.37\%, 22.09\% ,and 14.04\% on Wiki, DBLP and SemEval-Sci, respectively. 

Second, \textsc{CoL-Zero} demonstrates stronger domain adaptation capabilities than \textsc{HF-Zero} under the zero-shot setting. In the context of zero-shot learning, \textsc{CoL} (GPT-4) shows a 26.41\% improvement on $F1_e$ than \textsc{HF} (GPT-4) in DBLP, and for SemEval-Sci, \textsc{CoL} (GPT-4) achieves 14.42\% improvement than \textsc{GPT-4}. Although \textsc{HF-Zero} shows a better $F1_e$ than \textsc{CoL-Zero}, the increase in \textsc{HF-Zero}'s $F1_e$ over \textsc{CoL-Zero} is only 0.285\%, indicating that this marginal improvement is insufficient to prove that \textsc{HF-Zero} has superior domain adaptation capability compared to \textsc{CoL-Zero}.

These observations can be attributed to two reasons:  (1) \textsc{CoL} decomposes taxonomy induction into different sub-tasks, such as focusing on finding parent-child relationships within a given layer, which enables the model to learn how to do taxonomy induction domain transfer across different domains. (2) The Ensemble-based Ranking Filter effectively improves the model's precision without sacrificing recall. Compared to using the proposed filter to post-process the output results once, \textsc{CoL} allows the generated results to be corrected by the proposed filter at every iteration of building the taxonomy. This mechanism enables the model to perform self-correction on the output taxonomy based on the existing context.

\subsection{Investigating the Effects of Scalability on the \textsc{Chain-of-Layer} (RQ2)} 
\label{sec:exp_scalability}

\begin{figure}[t]
    \centering
    \includegraphics[width=1\linewidth]{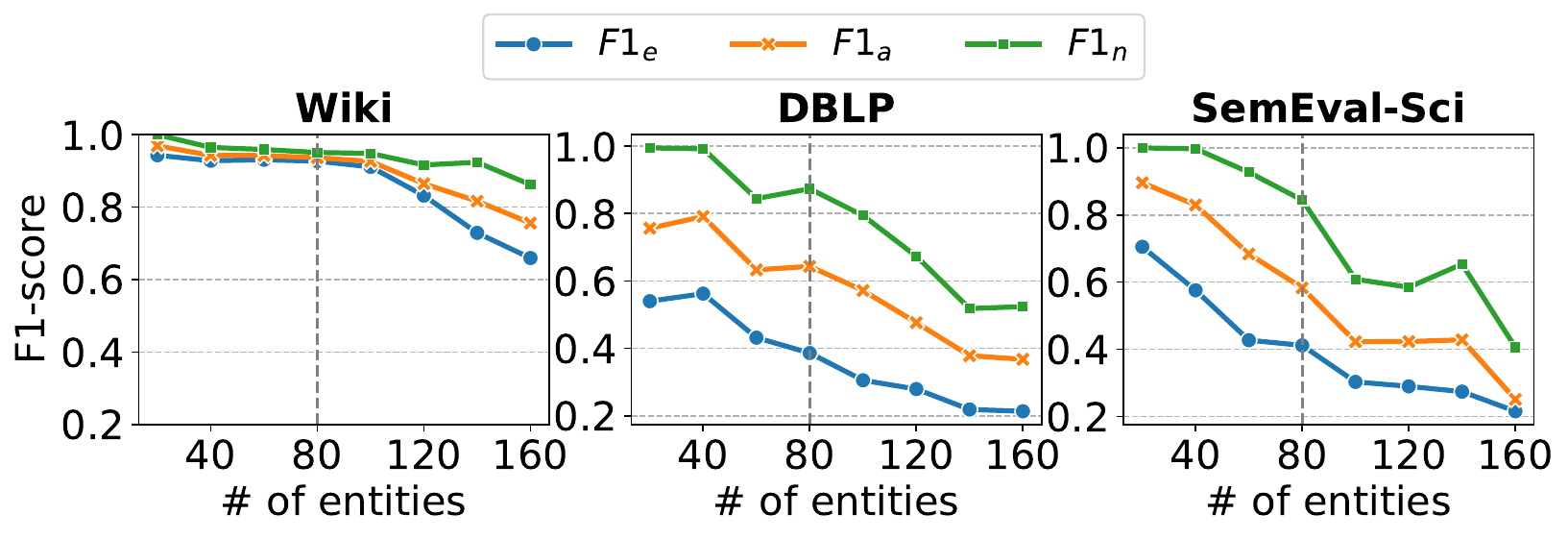}
    \vspace{-0.3in}
    \caption{Performance analysis of the \textsc{CoL} across varying scales and domains. It shows Edge, Ancestor, and Node F1-scores for Wiki, DBLP, and SemEval-Sci taxonomies, ranging from 20 to 160 entities. An inflection point at the 80-entity threshold across all metrics and domains, emphasizing the scalability limitations of \textsc{CoL}.}
    \label{fig:analysis_num_entity}
    \vspace{-0.25in}
\end{figure}

In this section, we empirically investigate the scalability of our proposed \textsc{Chain-of-Layer} framework across varying scales (number of entities in the given entity list), with particular emphasis on identifying a critical threshold below which proposed \textsc{CoL} demonstrates optimal performance. We conduct experiments on Wiki, DBLP, and SemEval-Sci taxonomies. For each taxonomy, we randomly select sub-taxonomies, using the root entity as the starting point. We chose sub-taxonomies of various sizes, specifically with 20, 40,  ..., 140, and 160 entities. To ensure the reliability of our results, we repeated the sampling process five times for each size, thereby generating five distinct sub-taxonomies for every specified number of entities. We not only report the trend of edge-level F1-score ($F1_e$ and $F1_a$) as it changes with variations in the size of the sub-taxonomy but also explore the trend of node-level F1-score ($F1_n$) on each dataset in Figure \ref{fig:analysis_num_entity}. Our observations are as follows.

First, as the scale of the taxonomy to be induced expands, both the edge-level F1-score and node-level F1-score of the proposed \textsc{CoL} framework exhibit a decline across all three benchmarks that in different domains. This correlation demonstrates that in the approaches that rely on prompting large language models, the increase in the number of entities significantly increases the complexity of the taxonomy induction task, leading to a relative performance decline even though the target taxonomy is in the same domain.

Secondly, in comparison to DBLP and SemEval-Sci,  \textsc{CoL} exhibits robustness on the Wiki taxonomy. Specifically, even when expanding the entity count in the taxonomy to 160,  \textsc{CoL} on  Wiki shows a decrease in $F1_e$ and $F1_a$ of 30.01\% and 22\%, respectively, compared to when the entity count is 20. In contrast, on DBLP, $F1_e$ and $F1_a$ decrease by 60.50\% and 51.41\%, respectively, and SemEval-Sci, $F1_e$ and $F1_a$  decrease by 69.49\% and 72.02\%, respectively. This differential performance decline indicates that LLMs have a stronger knowledge understanding in general domains than in specific domains, such as the scientific domain.

Last, we find that the node-level F1-score ($F1_n$) also decreases more drastically as the number of entities exceeds 80 on DBLP and SemEval-Sci. Notably, the $F1_n$ remains relatively high with 20-80 entities, it sharply declines beyond this point. These findings indicate that when the taxonomy scale exceeds a certain threshold (beyond 80 entities), LLMs struggle to strictly adhere to the rules mentioned in the instructions: \textit{using only the entities provided in the given entity set to carry out taxonomy induction.} This is also one of the significant reasons for the substantial decrease in the performance of \textsc{CoL} as the taxonomy scale increases. 

\begin{figure}[t]
    \centering
    \includegraphics[width=1\linewidth]{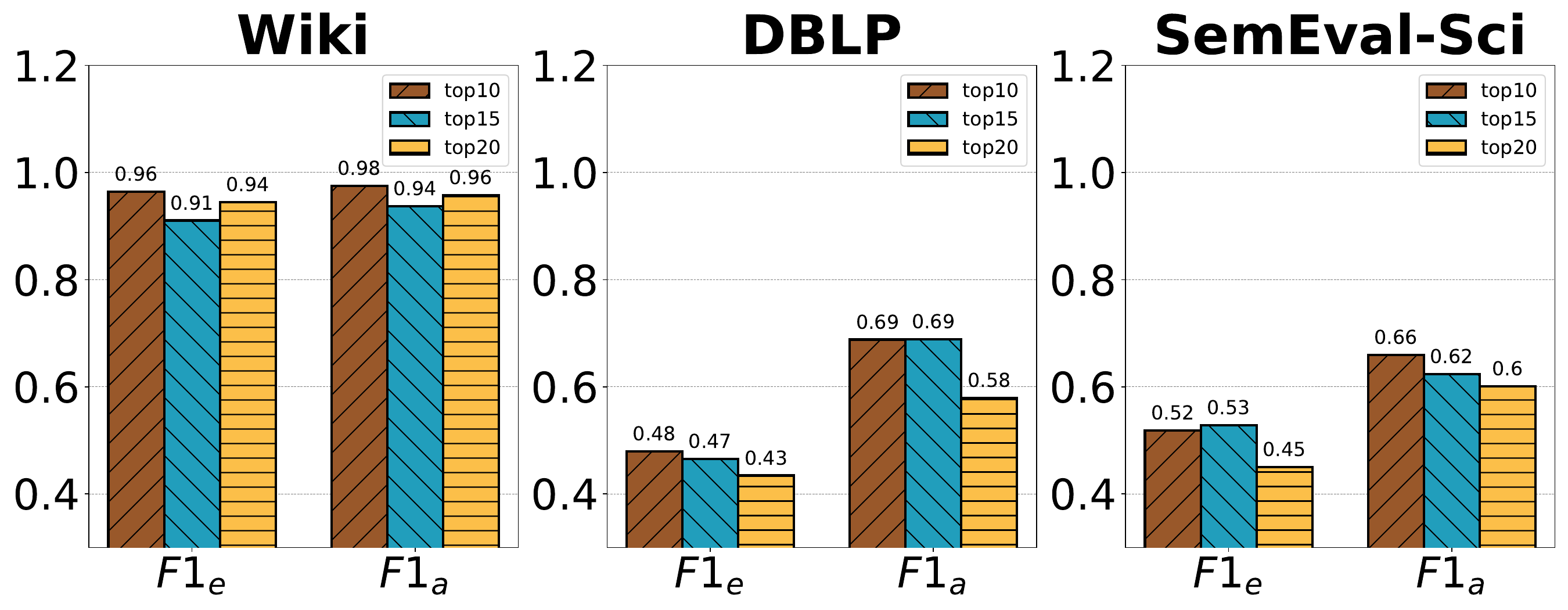}
    \vspace{-0.3in}
    \caption{Performance of ranking ranges in the Ensemble-based Ranking Filter for maintaining parent-child relationships. The Top-10 range shows the highest $F1_a$ scores across all datasets and the highest $F1_e$ scores in Wiki and DBLP. In SemEval-Sci, the Top-10 range achieves nearly the best $F1_e$ score, close to the Top-15 range.}
    \label{fig:analysis_topk}
    \vspace{-0.25in}
\end{figure}

\subsection{Investigating the Effects of Hyperparameters \& Ablation Study (RQ3)}




\subsubsection{The selection of the best ranking range}
To identify the best ranking range for maintaining the parent-child relationships produced by \textsc{CoL}, we evaluated the top-10, top-15, and top-20 rankings across three large-scale taxonomies. Given that the top-1 and top-5 rankings scored below 50\%, we consider them too stringent to accurately preserve the correct parent-child relationships. The results are presented in Figure~\ref{fig:analysis_topk}. The results are illustrated in Figure~\ref{fig:analysis_topk}. Our findings reveal that for both $F1_e$ and $F1_a$, the top-10 ranking consistently demonstrates the best or second-best performance across all three datasets. Consequently, we selected the top-10 ranking as the optimal range for preserving the parent-child relationship in the Ensemble-based Ranking Filter.
\subsubsection{Ablation Study}
We conducted an ablation study on the four benchmarks mentioned above to verify the effectiveness of two major modules: \textsc{Chain-of-Layer} prompting (\textsc{CoL}) and the Ensemble-based Ranking Filter (Filter) in the proposed framework. The experimental results are shown in Table~\ref{tab:Ablation_Study}. Our findings are as follows:

First, removing \textsc{CoL} or Ensemble-based Ranking Filter reduces performance on three three large-scale taxonomies (Wiki, DBLP, and SemEval-Sci). It proves that the incorporation of \textsc{CoL} and Ensemble-based Ranking Filter provide crucial self-correction, reducing hallucinated content.


Second, the most notable drop in recall and F1-score performance occurs when the \textsc{CoL} is removed. It indicates that utilizing the Ensemble-based Ranking Filter as a post-processing iteration for the generated taxonomy proves overly stringent in maintaining the parent-child relations, even when those relations are correct. On the DBLP dataset, the absence of \textsc{CoL}  results in a decrease of 65.7\% in $R_e$ and 46.8\% in $F1_e$, despite a 50.2\% improvement in $P_e$.

Third, removing the Ensemble-based Ranking Filter results in a decline in precision performance across all four benchmarks. This indicates that the proposed filter effectively preserves the accuracy of the parent-child relationship within the generated taxonomy.

Last, the introduction of \textsc{CoL} and the Ensemble-based Ranking Filter does not significantly impact the performance on WordNet. It is because WordNet's smaller scale allows models like \textsc{GPT-4 Turbo} to handle the task effectively without these enhancements.

\begin{table}[t]
    \centering
    \resizebox{1\linewidth}{!}{
    \begin{tabular}{lcccccccc}
         \toprule[1.5pt]
         \multirow{2}{*}{\textbf{Dataset}} & \multicolumn{2}{c}{\textbf{Configuration}} & \multicolumn{3}{c}{\textbf{Edge}} & \multicolumn{3}{c}{\textbf{Ancestor}}\\
         \cmidrule(lr){2-3} \cmidrule(lr){4-6} \cmidrule(lr){7-9}
         & \textbf{CoL} & \textbf{Filter} & $\textbf{P}_e$ & $\textbf{R}_e$ & $\textbf{F1}_e$ & $\textbf{P}_a$ & $\textbf{R}_a$ & $\textbf{F1}_a$\\
         \midrule
         & \xmark & \checkmark & 60.67 & 47.76 & 51.77 & 84.22 & 56.52 & 64.72 \\
         WordNet & \checkmark & \xmark & 59.12 & 58.41 & 58.76 & 90.11 & 74.31 & 80.77 \\
         & \checkmark & \checkmark & 59.57 & 57.10 & 57.73 & 90.60 & 73.07 & 79.62 \\
         \midrule
         Wiki & \xmark & \checkmark & 98.08 & 46.27 & 61.09 & 99.49 & 45.58 & 60.66 \\
         & \checkmark & \xmark & 98.58 & 93.08 & 95.73 & 99.51 & 93.96 & 96.63 \\
         & \checkmark & \checkmark & 97.92 & 94.99 & 96.43 & 99.17 & 95.99 & 97.54 \\
         \midrule
         DBLP & \xmark & \checkmark & 72.81 & 13.35 & 22.41 & 71.92 & 7.66 & 13.70 \\
         & \checkmark & \xmark & 48.47 & 38.87 & 42.14 & 69.29 & 59.94 & 63.44 \\
         & \checkmark & \checkmark & 55.07 & 44.27 & 47.96 & 79.95 & 63.06 & 68.82 \\
         \midrule
          SemEval-Sci & \xmark & \checkmark & 57.29 & 20.10 & 29.32 & 93.33 & 13.91 & 23.94 \\
          & \checkmark & \xmark & 54.22 & 49.75 & 51.86 & 86.29 & 53.74 & 65.97 \\
          & \checkmark & \checkmark & 59.60 & 46.03 & 51.59 & 91.23 & 48.16 & 62.69 \\
        \bottomrule[1.5pt]
    \end{tabular}
    }
    \caption{Ablation study of two major modules in the proposed framework: \textsc{Chain-of-Layer} prompting (\textsc{CoL}) and Ensemble-based Ranking Filter (Filter). All metrics are presented in percentages (\%). Configurations indicate whether \textsc{CoL} and the Ensemble-based Ranking Filter were employed.}
    \label{tab:Ablation_Study}
    \vspace{-0.3in}
\end{table}

\subsection{Case Study}

This section presents a case study to evaluate the strengths and weaknesses of our proposed methods alongside several baselines. We use samples from WordNet and provide outputs for \textsc{CoL}, \textsc{CoL-w/o-Filter}, \textsc{HF (GPT-4)}, and \textsc{HF (GPT-4)-Filter} in a 5-shot setting.

\subsubsection{CoL v.s. HF (GPT-4) / HF-Filter (GPT-4)}

\begin{figure*}[htbp]
    \centering
        \includegraphics[width=1\textwidth]{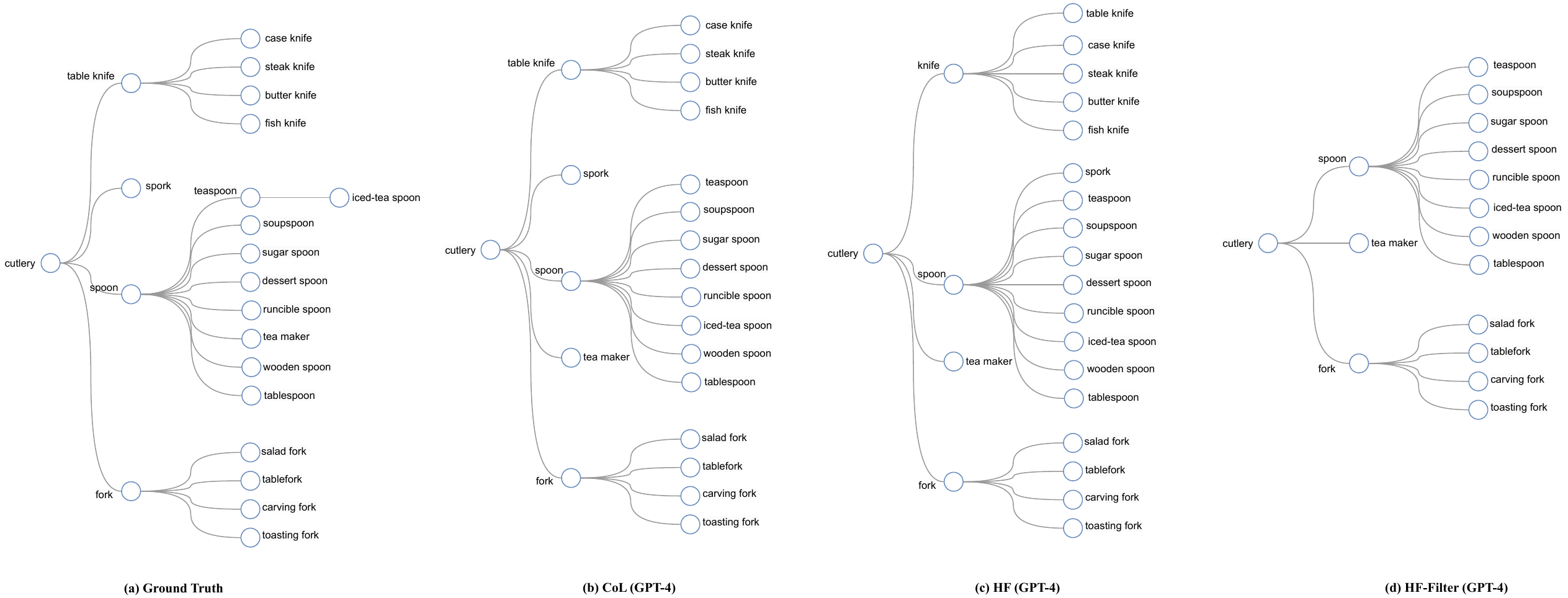}
    \caption{The taxonomies generated via \textsc{CoL},  HF (GPT-4) and HF-Filter (GPT-4). }
    \label{fig:case_1}
    \vspace{-0.1in}
\end{figure*}

By comparing the outputs of the \textsc{CoL} in Figure~\ref{fig:case_1}(b) and \textsc{HF (GPT-4)} in Figure~\ref{fig:case_1}(c)) / \textsc{HF (GPT-4)-Filter} in Figure~\ref{fig:case_1}(d), we demonstrate the effectiveness and importance of our \textsc{CoL} framework and the Ensemble-based Ranking Filter. Compared to the ground truth, the most noticeable issue with \textsc{HF (GPT-4)}'s output is that it hallucinates an entity \textit{knife} that wasn't in the given entity list and uses it to group all other entities that should belong to \textit{table knife}. This resulted in a lower edge F1 score for the generated taxonomy. 


When comparing the outputs of \textsc{HF (GPT-4)} and \textsc{HF (GPT-4)-Filter}, we observe that without a layer-by-layer decomposition approach like \textsc{CoL}, directly employing a filter degrades the quality of the induced taxonomy. Filtering \textsc{HF (GPT-4)}'s output results in the complete removal of the sub-taxonomy under \textit{knife}. This significantly lowers the node F1 score and edge F1 score of the generated taxonomy because filtered entities cannot be re-selected. These findings highlight the critical importance and synergistic effect of \textsc{CoL} and the Ensemble-based Ranking Filter.

\subsubsection{CoL v.s. CoL-w/o-Filter}

\begin{figure*}[htbp]
    \centering
        \includegraphics[width=1\textwidth]{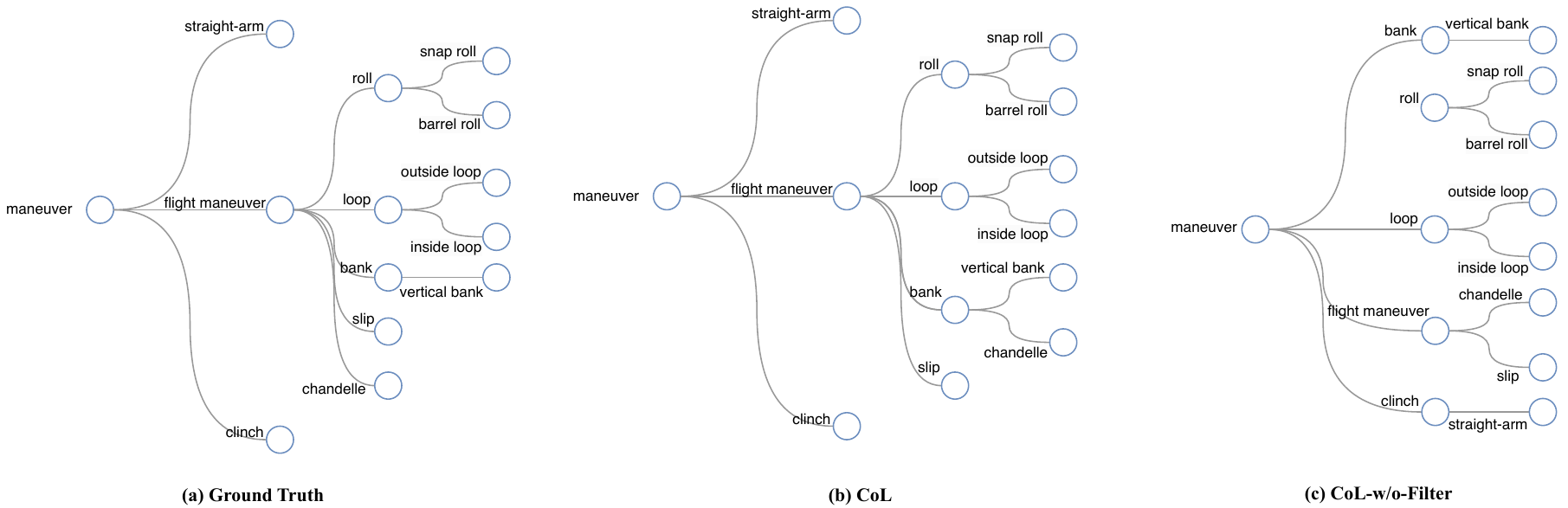}
    \caption{The taxonomies generated via \textsc{CoL} and CoL-w/o-Filter}
    \label{fig:case_2}
\end{figure*}

To illustrate the role of the Ensemble-based Ranking Filter, we compare the outputs of \textsc{CoL} in Figure~\ref{fig:case_2}(b) and \textsc{CoL-w/o-Filter} in Figure~\ref{fig:case_2}(c) against the ground truth in Figure~\ref{fig:case_2}(a). As shown, the taxonomy induced by \textsc{CoL} closely aligns with the ground truth, whereas \textsc{CoL-w/o-Filter} misclassifies ``\textit{roll}'', ``\textit{bank}'', and ``\textit{loop}'' as siblings of ``\textit{flight maneuvers}''. This demonstrates the effectiveness of the Ensemble-based Ranking Filter, which removes edges with lower ranks, such as ``\textit{flight maneuvers - roll}'' and re-adds ``\textit{roll}'' for selection in the next layer. This self-correcting process helps LLMs induce a more accurate taxonomy.

\section{Related Works}
\subsection{Taxonomy Induction}
The process of taxonomy induction typically includes identifying hypernyms (extracting potential parent-child relationships from text) and and organizing them hierarchically. In the initial stage,  embedding-based approaches~\cite{luu2016learning} and pattern-based methods~\cite{snow2004learning,nakashole2012patty,wu2012probase} were widely used. The second step was often viewed as graph optimization and solved by maximum spanning tree~\cite{bansal2014structured}. In this work, we focus on organizing a given entity set to a taxonomy. Bansal \emph{et al.}~\cite{bansal2014structured} approach this as a structured learning problem and employ belief propagation to integrate relational information between siblings. Mao \emph{et al.}~\cite{mao2018end} present an approach utilizing reinforcement learning to integrate the phases of hypernym identification and hypernym organization. Shang \emph{et al.}~\cite{shang2020taxonomy} utilize a graph neural network approach, demonstrating improvement in large-scale taxonomy induction using the SemEval-2016 Task 13 dataset~\cite{bordea2016semeval}. Chen \emph{et al.}~\cite{chen2021constructing} and Jain \emph{et al.}~\cite{jain2022distilling} utilize the pre-trained language model to approach taxonomy induction, treating it as sequence classification and sequence scoring tasks, respectively. Langlais and Guo~\cite{langlais2023rate} proposed an automatic taxonomy evaluation metric based on the pre-trained model. TaxonomyGPT~\cite{chen2023prompting} conducts taxonomy induction by leveraging the in-context learning capabilities of LLMs. The proposed \textsc{CoL} in this paper significantly reduces hallucination and improves structural accuracy by iterative prompting large language models.

\subsection{Extracting Knowledge from LLMs}
Research in extracting and investigating the stored knowledge in Large Language Models (LLMs) has become increasingly sophisticated, combining quantitative assessments with innovative extraction techniques. Prior works like LAMA~\cite{petroni-etal-2019-language}, TempLAMA~\cite{dhingra-etal-2022-time}, and MMLU~\cite{hendrycks2021measuring} have laid the groundwork by quantitatively measuring the factual and time-related knowledge within these models. Based on these, recent efforts have ventured into knowledge extraction, as seen in works that construct Knowledge Graphs (KGs) directly from LLM outputs. Specifically, methodologies like the one introduced in Crawling Robots~\cite{cohen-etal-2023-crawling} propose the extraction of named entities and relationships through a novel robot role-play setting, indicating a shift towards more interactive and dynamic extraction methodologies. Parallel to this, the adoption of structured, prompt-based queries has offered a pathway to not only retrieve but also systematically organize the knowledge embedded within LLMs, making it accessible and interpretable for human users \cite{liu-etal-2022-generated, yu2023generate}. This emerging body of work, including techniques that enhance training data with explicit knowledge recitation tasks \cite{sun2023recitationaugmented}, aims at not just understanding but also effectively leveraging the vast reservoir of information encapsulated in these advanced models, marking a significant leap forward in our quest to harness the full potential of LLMs for knowledge-based applications.

\section{Conclusion}
In this work, we introduce \textsc{Chain-of-Layer (CoL)}, a novel framework for taxonomy induction. By leveraging the hierarchical format instruction (\textsc{HF}) and incorporating an Ensemble-based Ranking Filter, \textsc{CoL} breaks down the task into selecting relevant candidates and gradually building the taxonomy from top to bottom and significantly reduces hallucination and improves structural accuracy. Extensive experimental results demonstrate that \textsc{CoL} outperforms various baselines, achieving state-of-the-art performance.  We envision \textsc{CoL} as a powerful framework to address the challenges of inducting accurate and coherent taxonomy from a set of entities.


\section*{Acknowledgment}
This work was supported by NSF IIS-2119531, IIS-2137396, IIS-2142827, IIS-2234058, CCF-1901059, and ONR N00014-22-1-2507.
\newpage
\bibliographystyle{ACM-Reference-Format}
\bibliography{main}



\end{document}